\documentclass[lettersize,journal]{IEEEtran}
\usepackage{amsmath,amsfonts}
\usepackage{algorithmic}
\usepackage{algorithm}
\usepackage{array}
\usepackage{textcomp}
\usepackage{stfloats}
\usepackage{url}
\usepackage{verbatim}
\usepackage{graphicx}
\usepackage{cite}
\usepackage{color}
\usepackage{multirow}
\usepackage[colorlinks = true,
            linkcolor = blue,
            urlcolor  = blue,
            citecolor = blue,
            anchorcolor = blue]{hyperref}
\begin{document}

\title{Asymmetric Cross-Scale Alignment for Text-Based Person Search}

\author{Zhong Ji, Junhua Hu, Deyin Liu, Lin Yuanbo Wu*, \textit{Senior Member, IEEE}, Ye Zhao
\thanks{This work was supported by the National Natural Science Foundation of China (NSFC) under Grant 62176178, U19A2073, 62002096, and the Natural Science Foundation of Tianjin under Grant 19JCYBJC16000.}
\thanks{Z. Ji and J. Hu are with the School of Electrical and Information Engineering, Tianjin University, Tianjin 300072, China. Email: \{jizhong,hujunhua\}@tju.edu.cn.}
\thanks{D. Liu is with Anhui Provincial Key Laboratory of Multimodal Cognitive Computation, School of Artificial Intelligence, Anhui University, Hefei 230039, China. Email: iedyzzu@outlook.com.}
\thanks{L. Y. Wu* (corresponding author) is with Department of Computer Science, Swansea University, SA1 8EN, United Kingdom. Email: l.y.wu@swansea.ac.uk.}
\thanks{Y. Zhao is with the Key Laboratory of Knowledge Engineering with Big Data, School of Computer Science and Information Engineering, Hefei University of Technology, Hefei 230009, China. Email: zhaoye@hfut.edu.cn.}
}

\markboth{Journal of \LaTeX\ Class Files,~Vol.~14, No.~8, August~2021}%
{Shell \MakeLowercase{\textit{et al.}}: A Sample Article Using IEEEtran.cls for IEEE Journals}


\maketitle

\begin{abstract}
Text-based person search (TBPS) is of significant importance in intelligent surveillance, which aims to retrieve pedestrian images with high semantic relevance to a given text description. This retrieval task is characterized with both modal heterogeneity and fine-grained matching. To implement this task, one needs to extract multi-scale features from both image and text domains, and then perform the cross-modal alignment. However, most existing approaches only consider the alignment confined at their individual scales, e.g., an image-sentence or a region-phrase scale. Such a strategy adopts the presumable alignment in feature extraction, while overlooking the cross-scale alignment, e.g., image-phrase. In this paper, we present a transformer-based model to extract multi-scale representations, and perform Asymmetric Cross-Scale Alignment (ACSA) to precisely align the two modalities. Specifically, ACSA consists of a global-level alignment module and an asymmetric cross-attention module, where the former aligns an image and texts on a global scale, and the latter applies the cross-attention mechanism to dynamically align the cross-modal entities in region/image-phrase scales. Extensive experiments on two benchmark datasets CUHK-PEDES and RSTPReid demonstrate the effectiveness of our approach. Codes are available at \href{url}{https://github.com/mul-hjh/ACSA}.
\end{abstract}


\section{Introduction}
\IEEEPARstart{T}{ext}-Based Person Search (TBPS) aims to retrieve the shots of a target person with high semantic relevance to a given text description. It has attracted increasing attention due to the wide applications in modern cities wherein a large number of monitoring devices are deployed. Compared to image query based approaches \cite{LAG-Net,CEAVA,VGG-D}, TBPS only requires a verbal description to query a target person. This setting is more practical in certain situations where the image query may not be accessible. Moreover, natural language can describe the target person more faithfully than alternative representations, such as attributes \cite{AMR, MTA-Net, CGCN}. Therefore, TBPS extends high practical value in real-world applications, such as multiple people tracking and person re-identification \cite{WU-TIP-Pseudo-Pair,WU-IJCAI-2022,LIU-TOMM}.

\begin{figure}[t] 
	\centering  
\includegraphics[height=4cm, width=9cm]{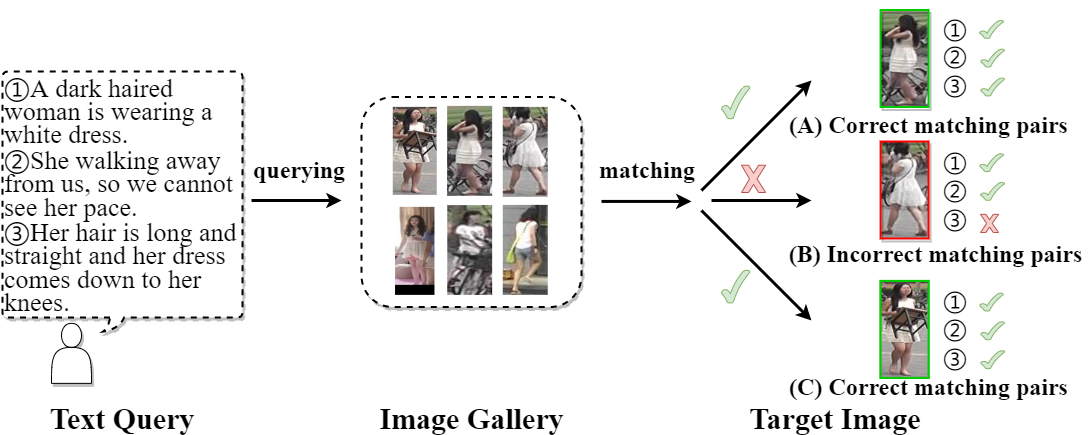}  
\caption{The cross-modal fine-grained nature of Text-Based Person Search (TBPS). Given a text query, TBPS retrieves the correct person images by matching the textual and image representations at both global and local scales with fine-grained details. See texts for details.}\label{fig:global-local}
\end{figure}

However, the task of TBPS has two open challenges. First, as a \textit{cross-modal} retrieval task, TBPS inherently encounters the modal heterogeneity problem, that is, the data distributions of different modalities are inconsistent \cite{LIU-TNNLS-Verbal-PersonNet,Chen-MM-2022}, and such modal heterogeneity makes it difficult to directly measure the correlation between visual and text representations. Second, TPBS is essentially a \textit{ fine-grained} visual search task, which requires the model to be effective in matching pedestrian images with high variations. In this case, a global-scale sentence may not be reliable cues to retrieve all correct images of the same identity. As shown in Fig. \ref{fig:global-local}, given a text query, the retrieved pedestrian images only matching the global sentence could be a different identity, while fine-grained matching with appearance details such as ``long hair" is more distinct to facilitate the matching.

\begin{figure}[!t] 
\centering  
\includegraphics[height=4cm, width=9cm]{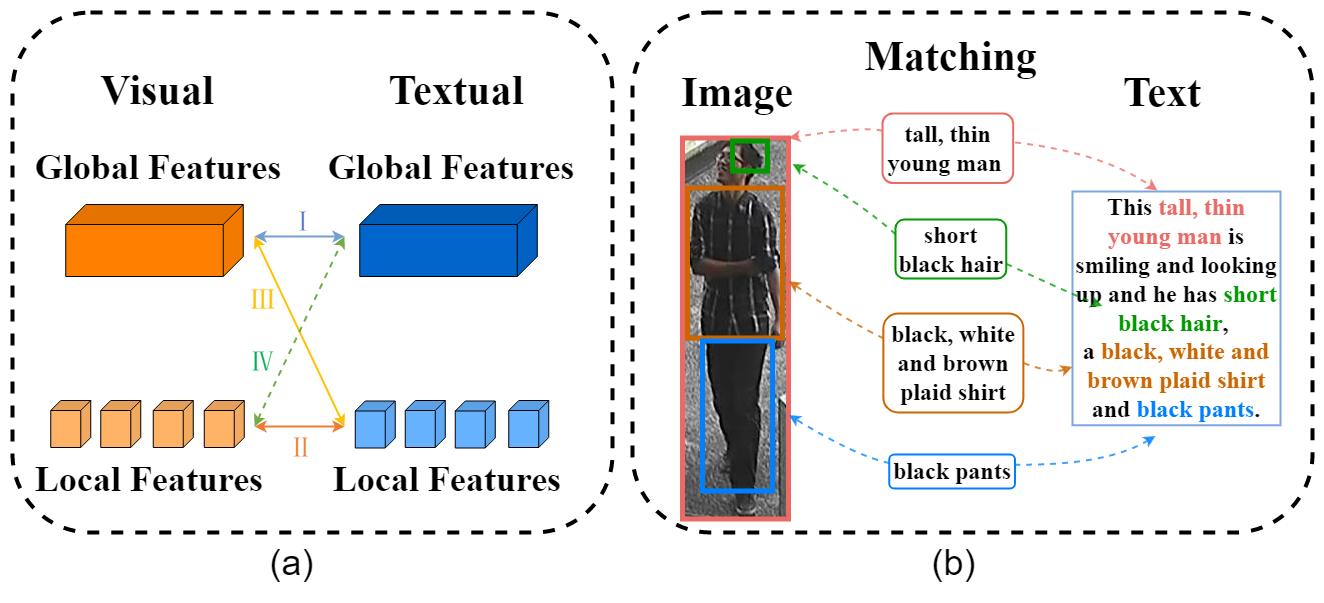}  
\caption{(a) Four cross-modal alignments between an image and texts. Here, I and II indicate the global (image-sentence) and local (region-phrase) alignments. III and IV represent the cross-scale alignments, i.e., image-phrase and region-sentence. (b) Natural language may use both overall description (e.g., tall, thin young man) and detailed phrases to describe the target person (e.g., short black hair, black pants).}
\label{fig:alignment}
\end{figure}

To address the modal heterogeneity problem, some studies have investigated the alignment between visual and text modalities. Early approaches only consider the global-level alignment \cite{GNA-RNN, PWM+ATH,DCMP}, which aligns the global visual with overall textual information, see relation \uppercase\expandafter{\romannumeral1} in Fig. \ref{fig:alignment} (a). Recall Fig. \ref{fig:global-local} that different pedestrians may look very similar in their overall appearance, and it is difficult to correctly distinguish them only from the global scale alignment. In this respect, local features are highly discriminative to facilitate the fine-grained matching. In this line, many attempts have been made to exploit the multi-granularity alignment \cite{MAA-Net,PMA,MIA,HGAN,RNN-HA}, which can include both global  and local visual-textual alignments, see relations \uppercase\expandafter{\romannumeral1} and \uppercase\expandafter{\romannumeral2} in Fig. \ref{fig:alignment} (a). In practice, local features are usually extracted by resorting to the auxiliary information, such as human poses \cite{PMA} and visual attributes \cite{ViTAA, CMAAM}. For example, Zheng \textit{et al}. \cite{HGAN} employed the hard-attention mechanism to select semantically relevant image regions and words/phrases, and performed the multi-granularity alignment on multiple levels, i.e., word-level, phrase-level and sentence-level. Recently, Gao \textit{et al}. \cite{NAFS} suggested that it is necessary to consider the additional cross-scale alignment, that is, adaptive alignment between different scales. For example, a word may correspond to a patch or the entire image, such as ``woman'', ``dress'' and ``slim'' are the overall description with regard to a person. Nevertheless, the words ``glasses'', ``bag'', and ``shoes'' can only describe some regions of an image. Thereby, they proposed the Non-local Alignment over Full-Scale representations (NAFS) \cite{NAFS}, which considers full-scale adaptive alignment. In other words, four alignment relations \uppercase\expandafter{\romannumeral1}, \uppercase\expandafter{\romannumeral2}, \uppercase\expandafter{\romannumeral3}, and \uppercase\expandafter{\romannumeral4} are fully employed in NAFS \cite{NAFS} (see Fig. \ref{fig:alignment} (a)).

However, such full-scale representation learning is based on the hypothesis that the cross-modal instances are symmetric in their information level, that is, each instance, e.g., the sentence can be associated with an entire image or a region, and vice versa. This may not hold true in TBPS. For instance, one may use a phrase to describe a whole image as outline but unlikely to use a long sentence to describe a region. As shown in Fig. \ref{fig:alignment} (b), a witness is likely to outline a person's body and gender, such as ``tall'', ``thin'', ``man''. Then, he tends to describe the details, such as ``short black hair'', ``black, white and brown plaid shirt'', ``black pants''. These information constitute the final complete text description. In fact, this process essentially includes three relations: overall description corresponds to a whole image, detailed texts correspond to a few regions of an image, and the complete texts correspond to the whole image again. Thus, it requires to extract multi-scale visual and textual representations, that is, global/local visual representations, global textual representations and phrase representations. Global visual features can be easily extracted using a pre-trained off-the-shelf model \cite{GNA-RNN}, while extracting local visual features is inexplicable, due to the presence of occlusion, background clutters in natural pedestrian images. To extract local features, some methods \cite{PMA,NAFS} applied object detection or additional branch networks to detect salient regions and then extract features. However, these methods yield high computational cost because of the external networks. Other methods \cite{MIA,AXM-Net} directly sliced the global visual representations horizontally into non-overlapping slices as local visual representations. This fashion is simple but may inadvertently divide the same part into different slices.

In this paper, we propose an Asymmetric Cross-Scale Alignment (ACSA) approach for TBPS. Specifically, we employ the Swin Transformer \cite{SwinTransformer} to extract the global visual representations, and then divide the global representation into four regions as local visual representations, namely head, upper body, lower body and feet. This partition strategy does not involve extra computational cost but can better preserve the salient body parts for fine-grained matching. In the text domain, we employ BERT \cite{BERT} to extract the global textual representations and local phrase representations. Both Swin Transformer \cite{SwinTransformer} and BERT \cite{BERT} are based on the self-attention mechanism, which can fully leverage the information within each modality.

We further propose an asymmetric cross-scale alignment module (ACSA) that performs three alignments, i.e., relations \uppercase\expandafter{\romannumeral1}, \uppercase\expandafter{\romannumeral2}, \uppercase\expandafter{\romannumeral3} in Fig. \ref{fig:alignment}. The proposed ACSA module consists of a global-level alignment and an asymmetric cross attention. The former is to perform the global image-text alignment, i.e., relation \uppercase\expandafter{\romannumeral1}, and the latter is to adaptively align the image/regions with phrases, i.e., relation \uppercase\expandafter{\romannumeral2} and relation \uppercase\expandafter{\romannumeral3}. Compared to the multi-granularity alignment based methods \cite{MAA-Net,PMA,MIA,HGAN,RNN-HA}, the proposed ACSA performs cross-scale image-phrase alignment, and the phrase is adaptively aligned with an entire image or regions. In other words, a phrase may correspond to either a region or the whole image. This alignment can be automatically learned through the cross-attention mechanism in the network, rather than restricting the alignment at a certain scale, e.g., a global or local scale. Compared to a recent adaptive full-scale alignment method \cite{NAFS}, the proposed ACSA does \textit{not} perform the region-text alignment, i.e., relation \uppercase\expandafter{\romannumeral4} in Fig. \ref{fig:alignment}. This alignment is empirically demonstrated to be unnecessary for the task of TBPS. In other words, a full-scale alignment can cause the over-matching, while not contribute to the matching performance. Please refer to our experimental results in Table \ref{tab:5}.

The main contributions of this paper are summarized below.
\begin{itemize}
\item We propose an Asymmetric Cross-Scale Alignment approach, which exploits three effective alignments, namely image-text, region-phrase, and image-phrase alignments, to improve the performance of TBPS.
\item We develop a transformer-based framework to extract multi-scale feature representations, including the global and local representations for both image and text domains. With such multi-scale features, the cross-modal matching is performed with the proposed asymmetric cross-attention mechanism.
\item Our approach achieves state-of-the-art performance on two public datasets. Extensive ablation studies and visualization demonstrate the effectiveness of our approach.
\end{itemize}
\section{Related works} \label{sec:related}

\subsection{Text-Based Person Search}

Since Li \textit{et al}. \cite{GNA-RNN} first established the task of TBPS and released a large-scale dataset called CUHK-PEDES. Since then, TBPS has became a popular topic in intelligent surveillance. It can search relevant person images using natural language as the query instead of using images or attributes \cite{CEAVA,VGG-D,DCC-s,MTA-Net,WU-Video-Re-ID}, and thus has shown high practical value in real-world applications \cite{Wang-survey}. Most existing studies employ the following steps: (1) Applying CNNs or RNNs to extract the respective visual and textual features from images and texts; (2) Projecting these cross-modal features into a common latent feature space, followed by alignment; (3) Calculating the similarity of the image-text pair. Roughly, existing methods, based on their different focuses, can be categorized into three streams: feature representation approaches, cross-modal alignment approaches and approaches focusing on loss functions of similarities.

The first group focuses on the feature representations. For the visual modality, CNNs are the most popular backbones, such as VGG-Net \cite{GNA-RNN,PWM+ATH}, MobileNet \cite{DCMP,MCCL}, and ResNet \cite{TIMAM,GLILA}. As for the textual modality, early studies usually employed RNNs \cite{GNA-RNN,DCMP}, while some methods employed CNNs \cite{MAA-Net,Dual-Path}. For example, Zhang \textit{et al}. \cite{DCMP} tokenized the sentence and split it into words, and then sequentially processed them with a bi-directional LSTM, which is a variant of RNNs. Zheng \textit{et al}. \cite{Dual-Path} proposed a dual-path CNN to learn the image and text representations. Since TBPS is a fine-grained search task, local discriminative features are imperative. Therefore, attention mechanism can be leveraged, which seeks to boost local information or mitigate the noisy interference residing in global features. Li \textit{et al}. \cite{GNA-RNN} proposed a GNA-RNN model with the gated neural attention mechanism, taking into account that different words have different importance. Ji \textit{et al}. \cite{MAA-Net} applied attention mechanism to learn discriminative representations, and offered an accurate guidance to a common space. Different from the existing works, we employ Swin Transformer \cite{SwinTransformer} and BERT \cite{BERT} to extract visual and textual representations, respectively. They both are based on self-attention mechanism.

The second group investigates the alignment of visual and textual modalities. For example, Zhang \textit{et al}. \cite{DCMP} learned the global representations of images and texts, and then aligned the images with sentences, without involving local alignment. Afterwards, multi-scale alignment received great attention, in which the local-level alignment is employed as an important supplement to the global-level alignment. Jing \textit{et al}. \cite{PMA} utilized pose information to guide visual feature extraction, thereby learning latent semantic alignment between visual part and textual noun phrase. Niu \textit{et al}. \cite{MIA} proposed a multi-granularity image-text alignment model. Particularly, they first extracted the features of image parts and noun phrases as local representations, and then performed multi-granularity alignment, i.e., global-global alignment, global-local alignment, and local-local alignment. Zheng \textit{et al}. \cite{HGAN} proposed a hierarchical Gumbel attention network, which adaptively selected the semantically relevant image regions and words/phrases for precise alignment. Their matching strategy includes three granularities, i.e., word-level, phrase-level, and sentence-level. Recently, cross-scale alignment was developed to indicate that the alignments between different scales are also beneficial. Gao \textit{et al}. \cite{NAFS} proposed non-local alignment over full-scale representations. They designed a novel staircase CNN network and a locality-constrained BERT model to extract multi-scale visual and textual representations, and applied a contextual non-local attention mechanism to align the learned representations across different scales adaptively. Different from them, as discussed above, we extract multi-scale representations for performing asymmetric cross-scale alignment. We also propose a partition strategy to obtain local visual representations without computational cost.

The third group aims to develop different loss functions for similarities. For instance, Zheng \textit{et al}. \cite{Dual-Path} proposed the instance loss that explicitly considers the intra-modal data distribution, and each image/text group is viewed as a class. Zhang \textit{et al}. \cite{DCMP} proposed a cross-modal projection matching (CMPM) loss and a cross-modal projection classification (CMPC) loss for learning a discriminative image-text embedding. Both CMPM and CMPC losses are effective in global-level alignment. Besides, in this paper we design an asymmetric cross-scale alignment loss based on KL divergence for cross-scale alignment.

\subsection{Transformer}

Transformer was first proposed in \cite{Transformer} for addressing machine translation tasks. Instead of using recurrent formulation, it only employs the self-attention mechanism, and thus Transformer has a better parallel ability and yet alleviates the problem of long-distance dependence of texts. Based on Transformer, Devlin \textit{et al}. \cite{BERT} proposed a pre-training BERT model, which has good generalization ability and achieve promising progress on multiple NLP tasks.

With the resurgence a series of Transformer models in NLP \cite{XLNet,RoBERTa,GPT-3}, their applications in computer vision have also attracted increasing attention. Dosovitskiy \textit{et al}. \cite{ViT} proposed a seminal Vision Transformer model (ViT), which interprets an image as a sequence of patches and processes them with a standard Transformer encoder as that in NLP. However, there are two drawbacks in ViT. First, too many patches in high-resolution images will cause high computational complexity. Second, the fixed split patch mode is difficult to adapt to the problem of variable scale in computer vision. To address the above challenges, Liu \textit{et al}. \cite{SwinTransformer} proposed Swin Transformer, a hierarchical Transformer whose representation is calculated with shift windows. This way of grouping patches significantly reduces the computational complexity. In addition, by gradually merging patches, its view of field is gradually increased. This renders it more suitable for computer vision tasks.

Recently, Transformer has achieved state-of-the-art performance on multiple computer vision tasks \cite{DETR,SETR,FocalTransformer}. Chen \textit{et al}. \cite{IPT} applied it to low-level computer vision task, and achieved state-of-the-art performance on several tasks like super-resolution, denoising, and de-raining. He \textit{et al}. applied ViT to the Re-ID task \cite{TransReID}, in which they employed a sliding window to generate overlapping patches as the input of ViT to maintain the local neighbor structure information of the patch. Liang \textit{et al}. \cite{SwinIR} employed Swin Transformer in image restoration task, where multiple Residual Swin Transformer Blocks (RSTB) is developed to extract deep features, and each RSTB is composed of multiple Swin Transformer layers and residual connection. The shift window mechanism in RSTB can perform long-distance dependency modeling.

\section{Our Approach} \label{sec: approach}

\begin{figure*}[!t]
\centering
\includegraphics[height=9cm,width=17cm]{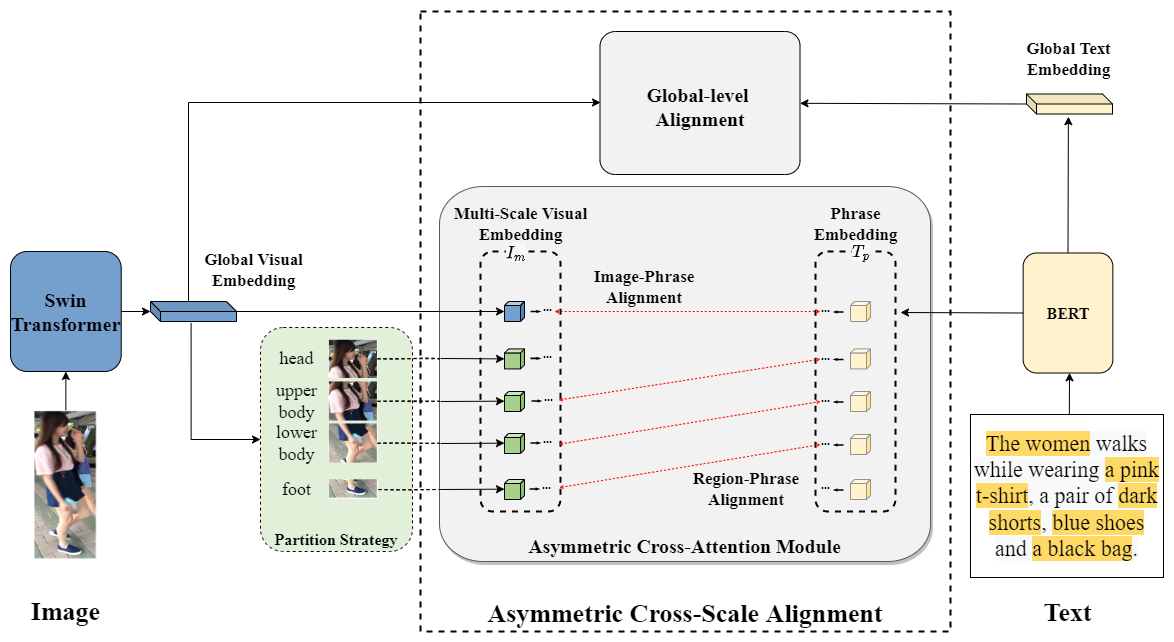}  
\caption{The framework of our approach consists of three parts: multi-scale visual and textual representation extraction and the proposed Asymmetric Cross-Scale Alignment module (ACSA). ACSA includes a global-level alignment module for image-text alignment, and an asymmetric cross-attention module for adaptive region-phrase and image-phrase alignments.}
\label{framework}
\end{figure*}

Text-Based Person Search (TBPS) can be regarded as a fine-grained cross-modal retrieval task which simultaneously deals with multi-granularity alignment and modal heterogeneity. The major challenge is to extract finer features from texts and images at multiple levels, and appropriately align these instances across modalities. To this end, we design a framework on the basis of Swin Transformer \cite{SwinTransformer} and BERT \cite{BERT} to extract multi-scale representations from images and texts, and perform the Asymmetric Cross-Scale Alignment (ACSA), i.e., global-level image-text, local-level region-phrase, and cross-scale image-phrase alignments. As shown in Fig. \ref{framework}, we use the output of Swin Transformer \cite{SwinTransformer} as the global image representation, and then divide this representation into head, upper body, lower body, and foot regions, which form the local image representations. BERT \cite{BERT} is employed as an encoder to extract the multi-scale representations of a given text, i.e., a global text representation and a set of noun phrase representations. In the following, we first detail the feature extraction from images and texts (section \ref{ssec:visual} and \ref{ssec:textual}). Then, we present the proposed Asymmetric Cross-Scale Alignment (section \ref{ssec:ACSA}).

\subsection{Visual Representations} \label{ssec:visual}
We extract the representations from images in both global and local levels. The global representation integrates all the information of a person. The local features provide fine-grained details with respect to body parts and patterns, etc. We argue that both global and local features should be leveraged to faithfully describe the identities in images.

\textbf{Global Representations.} Given an image \textit{I}, we aim to encode it into a vector $I_{g} \in R^{1 * d}$, where \textit{d} is the dimensionality of the feature vector. We adopt Swin Transformer \cite{SwinTransformer} as the backbone feature extractor due to its capability of extracting hierarchical features at both global and local levels. Specifically, we first resize the image \textit{I} to 224x224, then we divide it into patches to fit the Swin Transformer \cite{SwinTransformer}. We apply the pre-trained Swin Transformer model \cite{SwinTransformer}, and fine-tune it on our training data. Since Swin Transformer \cite{SwinTransformer} gradually increases its field of view as the network deepens, we use the feature output of Stage 4 (after the global pooling) as the global visual representation, i.e., $I_{g}$. 

\textbf{Local Representations.} We divide the pedestrian image into several regions according to its characteristics, i.e., the head (e.g., cap, hairstyle, glasses), the upper body (e.g., jacket, backpack), the lower body (e.g., pants, handbag), and the foot regions (e.g., shoes), we extract visual features from these four regions to form the local representations. Specifically, we divide the global visual representation into six parts horizontally, then employ the first and the second parts as the head representation, the second and third parts as the upper body representation, the fourth and fifth parts as the lower body representation, and the sixth part as the foot region representation. The head and upper body regions are partially overlapping, because some components may cross the two regions, such as long hair, scarf, etc. We additionally apply a fully connected layer to adjust all representations into the same dimension. Finally, we concatenate the four region embeddings to be the local image embeddings, denoted as $I_r =\left[I_{head}, I_{upper}, I_{lower}, I_{foot}\right] \in R^{k*d}$, where $k=4$ and \textit{d} is the dimensionality of the embedding.

\subsection{Textual Representations} \label{ssec:textual}
We employ the Bidirectional Encoder Representation from Transformers (BERT) \cite{BERT} to extract textual features. The self-attention mechanism in BERT \cite{BERT} can fully make use of the contextual relationship between words, allowing each word to establish a connection with any other word. 

\textbf{Global Representations.} To fit the input requirement of BERT \cite{BERT}, we split the texts into words and tokenize each word to be a token, then we insert [CLS] and [SEP] tokens at the beginning and end, respectively. We set the maximum number of tokens to \textit{L}, if there are less than \textit{L}, we fill them with zeros, and if there are more than \textit{L} tokens, we take the first \textit{L} tokens. Then we input the processed texts into the pre-trained BERT \cite{BERT}, we take the [CLS] as the global text representation, denoted as $T_{g} \in R^{1*d}$.

\textbf{Local Representations.} Li et.al \cite{GNA-RNN} suggest that nouns have more discriminant information. Thus, we extract noun phrases from texts as local text representations. Specifically, we use the TextBlob tool \cite{NLTK} to extract $M$ noun phrases from every text, and encode them into feature vectors in a similar way to the above textual encoding. The final local text representations are represented as $T_{p} =\left[t_1, t_2, \ldots, t_M\right] \in R^{M*d}$.

\subsection{Asymmetric Cross-Scale Alignment}\label{ssec:ACSA}

Our observation is that the region-text alignment does not hold true in TBPS, while other forms of alignment in terms of image-text, region-phrase and image-phrase are beneficial to the task. Therefore, we propose an Asymmetric Cross-Scale Alignment (ACSA) approach. The proposed ACSA consists of two major components: a global-level image-text alignment, and region-phrase/image-phrase alignments based on an asymmetric cross-scale attention module.

\textbf{Global Alignment.} We employ the Cross-Modal Projection Matching (CMPM) loss \cite{DCMP} and the Cross-Modal Projection Classification (CMPC) loss \cite{DCMP} for the global-level matching, which are demonstrated to be effective in learning cross-modal visual and textual representations. The CMPM loss associates the representations of different modalities by integrating the cross-modal projections into the KL divergence. Moreover, the CMPC loss applies identity-level annotations for cross-modal projection classification, so as to increase the differences of features for inter-class samples and enhance the compactness of features for intra-class samples. More details about the two losses can be found in \cite{DCMP}.

\textbf{Asymmetric Cross-Attention Module.} Inspired by Lee \textit{et al}. \cite{SCAN}, we propose an Asymmetric Cross-Attention Module (ACAM) to perform the region-phrase alignment and the image-phrase alignment. As shown in Fig. \ref{framework}, ACAM takes two inputs: a set of multi-scale visual embeddings $I_m=\left[v_{0}, v_{1}, \ldots, v_{k}\right] \in R^{(k+1) * d}$, which are obtained by concatenating the global image embedding $I_{g}$ and the region embeddings $I_{r}$, and a set of noun phrase embeddings $T_{p}=\left[t_{1}, t_{2}, \ldots, t_{M}\right] \in R^{M*d}$. The output of ACAM is a similarity score, which measures the similarity of an image-text pair by using $I_m$ and $T_p$.

Intuitively, ACAM attempts to align image-phrase and region-phrase instances in the sense that a noun phrase may correspond to the whole image or the image partially. However, we do not perform the region-text alignment because it is unlikely that all textual descriptions are trying to describe a specific part of an image. Therefore, the alignment relationship between texts and images is asymmetric. More specifically, in ACAM, we use visual entities and noun phrases as contexts for each other, while paying different attention to them to obtain the weighted visual and textual representations, and then calculate the similarity for the image-text pair. In the following, we detail the two directional cross-attention based alignment nested in ACAM.

\begin{figure}[t]
\centering  
\includegraphics[height=6.5cm, width=8cm]{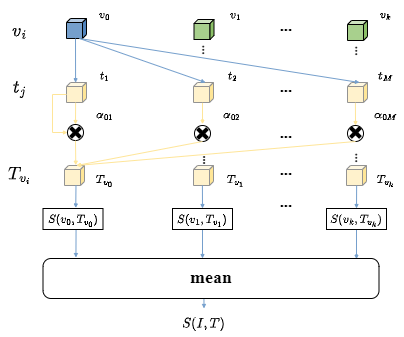}  
\caption{Illustration of the image-text cross-attention module. We first calculate the similarity between $v_i$ ($i=0,\dots, k$) and each phrase $t_{j}$ ($j=1,\dots,M$), and thus obtain the corresponding weighted text representation $T_{v_{i}}$. Herein, $k$ does not have to equal to $M$. Then we calculate the similarity between $v_i$ and $T_{v_i}$ as $S(v_i, T_{v_i})$. The final similarity of the image-text pair $S(I,T)$ is computed by averaging the values of $\sum_{i=0}^k S(v_i, T_{v_i})$.}
	\label{cross-attention}
\end{figure}

\textbf{Image-Text Cross-Attention.} For a specific visual entity \textit{i} (the entity may be a region or an image), we calculate the similarity between its embedding $v_{i}$ and every noun phrase embedding $t_j$ as follows:
\begin{equation} \label{eq1}
	s_{i j}=\frac{v_{i}^{T} t_{j}}{\left\|v_{i}\right\|\left\|t_{j}\right\|}, i \in[0, k], j \in[1, M].
\end{equation}
Here, $s_{i j}$ represents the similarity between the \textit{i}-th visual entity and the \textit{j}-th phrase. We normalize $s_{i j}$ as
\begin{equation} \label{eq2}
	\tilde{s}_{i j}=\frac{\left[s_{i j}\right]_{+}}{\sqrt{\sum_{i=0}^{k}\left[s_{i j}\right]_{+}^{2}}},
\end{equation}
where $\left[s_{i j}\right]_{+}=\max \left(s_{i j}, 0\right)$. We further employ the attention mechanism to obtain the weighted text representations with respect to the visual entity \textit{i}:
\begin{equation} \label{eq4}
	T_{v_{i}}=\sum_{j=1}^{M} \alpha_{i j} t_{j},
\end{equation}
where $\alpha_{i j}$ is the attention weight that can be calculated as
\begin{equation} \label{eq5}
	\alpha_{i j}=\frac{\exp \left(\lambda_{1} \tilde{s}_{i j}\right)}{\sum_{j=1}^{M} \exp \left(\lambda_{1} \tilde{s}_{i j}\right)}.
\end{equation}
In Eq. \eqref{eq5}, the parameter $\lambda_{1}$ is the inversed temperature of the softmax function. Then, the similarity between the visual representation $v_{i}$ and the corresponding weighted text representation $T_{v_{i}}$ is computed as:
\begin{equation} \label{eq6}
	S\left(v_{i}, T_{v_{i}}\right)=\frac{v_{i}^{T} T_{v_{i}}}{\left\|v_{i}\right\|\left\|T_{v_{i}}\right\|}.
\end{equation}
By averaging all $S\left(v_{i}, T_{v_{i}}\right)$, we obtain the final similarity for an image-text pair:
\begin{equation} \label{eq7}
	S(I, T)=\frac{\sum_{i=0}^{k} S\left(v_{i}, T_{v_{i}}\right)}{k+1}.
\end{equation}

Fig. \ref{cross-attention} shows the intuition of the above image-text cross attention mechanism. If the visual entity \textit{i} is not described in the text, the similarity value between $v_{i}$ and $T_{v_{i}}$ will be small. This is because when we calculate $T_{v_{i}}$, the visual vector $v_{i}$ and each phrase embedding $t_j$ is assumed to have a uniform similarity distribution, which results in an unweighted attention to a specific phrase (For example, $\alpha_{i1}=\alpha_{i2}= \ldots=\alpha_{iM}=1/M)$. If a phrase $t_j$ is important to describe a subject,  paying no attention to $t_j$ will cause a small similarity between $v_{i}$ and $T_{v_{i}}$ in Eq. \eqref{eq6}. We particularly reflect the importance of each phrase associated with the image, and formulate this rationale into the embedding based similarity function. Therefore, we measure the importance of the visual entity \textit{i} with respect to the texts by calculating the similarity between $v_{i}$ and $T_{v_{i}}$.

\textbf{Text-Image Cross-Attention.} Similarly, for the noun phrase \textit{j}, we calculate the similarity between its embedding $t_{j}$ and all visual entities as follows:
\begin{equation} \label{eq8}
	s_{i j}^{\prime}=\frac{v_{i}^{T} t_{j}}{\left\|v_{i}\right\|\left\|t_{j}\right\|}, i \in[0, k], j \in[1, M].
\end{equation}
Here, $s_{i j}^{\prime}$ represents the similarity between the \textit{i}-th visual entity and the \textit{j}-th phrase. We normalize it as
\begin{equation} \label{eq9}
	\tilde{s_{i j}^{\prime}}=\frac{\left[s_{i j}^{\prime}\right]_{+}}{\sqrt{\sum_{j=1}^{M}\left[s_{i j}^{\prime}\right]_{+}^{2}}},
\end{equation}
where $\left[s_{i j}^{\prime}\right]_{+}=\max \left(s_{i j}^{\prime}, 0\right)$. With these similarity values, we employ the attention mechanism to obtain a weighted visual representation related to the phrase \textit{i}:
\begin{equation} \label{eq11}
	V_{t_{j}}=\sum_{j=1}^{M} \alpha_{i j}^{\prime} t_{j},
\end{equation}
where
$\alpha_{i j}^{\prime}=\frac{\exp \left(\lambda_{1}^{\prime} \tilde{s}_{i j}^{\prime}\right)}{\sum_{j=1}^{M} \exp \left(\lambda_{1}^{\prime} \tilde{s}_{i j}^{\prime}\right)}$ and $\lambda_{1}^{\prime}$ is the inversed temperature of the SoftMax function. Then we calculate the similarity between the phrase embedding $t_{j}$ and the corresponding weighted visual representation $V_{t_{j}}$ as follows:
\begin{equation} \label{eq13}
	S\left(t_{j}, V_{t_{j}}\right)=\frac{t_{j}^{T} V_{t_{j}}}{\left\|t_{j}\right\|\left\|V_{t_{j}}\right\|}.
\end{equation}
By averaging all $S\left(t_{j}, V_{t_{j}}\right)$, we obtain the similarity of a text-image pair:
\begin{equation} \label{eq14}
	S^{\prime}(I, T)=\frac{\sum_{j=1}^{M} S\left(t_{j}, V_{t_{j}}\right)}{M}.
\end{equation}

\textbf{Asymmetric Cross-Scale Alignment Loss.} We apply KL divergence to associate the representations across different modalities for cross-scale matching. Given a mini-batch with \textit{N} image-text pairs, for each image $x_{a}$ the image-text pair is constructed as $\left\{\left(x_{a}, z_{b}\right), y_{a, b}\right\}_{a=1}^{N}$, where $y_{a, b}=1$ means that $\left(x_{a}, z_{b}\right)$ is a positive pair, while $y_{a, b}=0$ indicates $\left(x_{a}, z_{b}\right)$ is a negative pair. The probability of matching $x_{a}$ to $z_{b}$ is defined as
\begin{equation} \label{eq15}
	p_{a, b}=\frac{S\left(x_{a}, z_{b}\right)}{\sum_{a=1}^{N} S\left(x_{a}, z_{b}\right)},
\end{equation}
where $S\left(x_{a}, z_{b}\right)$ is the similarity between $x_{a}$ and $z_{b}$.

Since each identity can be associated with multiple images and multiple texts, this may incur a multi-matching \textit{z} for $x_{a}$ in a mini-batch, so we normalize the true matching probability as
\begin{equation} \label{eq16}
	q_{a, b}=\frac{y_{a, b}}{\sum_{m=1}^{N} y_{a, m}}.
\end{equation}
Then we apply KL divergence to measure the distance between the actual distribution $p_{a,b}$ and the true distribution $q_{a,b}$:
\begin{equation} \label{eq17}
	\mathcal{L}_{a}=\sum_{b=1}^{N} p_{a, b} \log \frac{p_{a, b}}{q_{a, b}+\varepsilon},
\end{equation}
where $\varepsilon$ is a non-negligible value to avoid numerical problems. We compute the matching loss from images to texts in a mini-batch as follows:
\begin{equation} \label{eq18}
	\mathcal{L}_{i 2 t}=\frac{1}{N} \sum_{a=1}^{N} \mathcal{L}_{a}.
\end{equation}
Similarly, we compute the matching loss from texts to images in the mini-batch as follows:
\begin{equation} \label{eq19}
	\mathcal{L}_{t 2 i}=\frac{1}{N} \sum_{b=1}^{N} \mathcal{L}_{b}.
\end{equation}
The matching loss is calculated as $\mathcal{L}_{ACSA}=\mathcal{L}_{i 2 t}+\mathcal{L}_{t 2 i}$.

\textbf{Objective Function.}
The final objective function is formulated as
\begin{equation} \label{eq21}
	\mathcal{L}_{\text {total }}=\mathcal{L}_{\mathrm{cmpm}}+\mu \mathcal{L}_{\mathrm{cmpc}}+\gamma \mathcal{L}_{\mathrm{ACSA}},
\end{equation}
where $\mu$ and $\gamma$ are hyperparameters to control the importance of different loss functions. We conduct a study on hyperparameters, and we set $\mu=4$ and $\gamma=0.1$ as default.
\section{Experiments}\label{sec:exp}

\subsection{Datasets and Evaluation Protocol}

\textbf{CUHK-PEDES.} CUHK-PEDES dataset \cite{GNA-RNN} is the first large public dataset for TBPS, which contains 40,206 images of 13,003 pedestrians, and each image has two text descriptions. The average length of text descriptions is 23.5 words. Following \cite{GNA-RNN}, we divided the dataset as follows. The training set has 11,003 pedestrians, 34,054 images and 68,126 captions. The validation set has 1,000 pedestrians, 3,078 images and 6,158 captions. The test set has 1,000 pedestrians, 3,074 images and 6,156 captions. 

\textbf{RSTPReid.} RSTPReid dataset is a new dataset constructed by Zhu \textit{et al}. \cite{DSSL} based on the MSMT17 dataset. This dataset is more challenging than CUHK-PEDES dataset. RSTPReid contains 20,505 images of 4,101 pedestrians. Each pedestrian has five images with each image corresponding to two text descriptions. Following DSSL \cite{DSSL}, we divided the dataset into the training set (3,701 images), validation set (200 images) and test set (200 images).

\textbf{Evaluation Protocol.} We adopted the widely used top-$k$ accuracy as the retrieval criterion \cite{GNA-RNN}, which ranks all gallery images according to their similarity with respect to the text query. If the correct pedestrian image is found in the first $k$ ranked images, the retrieval is considered to be successful.

\subsection{Implementation Details}
All images are resized to 224x224. We employed the pre-trained Swin Transformer Tiny \cite{SwinTransformer} as the visual backbone to extract features from images. The BERT \cite{BERT} pre-trained on the CUHK-PEDES dataset is used as the text backbone. The maximum number of tokens is set to 100. The embedding dimension is set to 768. The inverse temperature of softmax is set to 20.0. We employed the AdamW optimizer \cite{Adam} for 30 epochs with the Cosine decay learning rate scheduler and 5 epochs of linear warm-up. The batch size is set to 16. The initial learning rate is 0.0001, and the minimum learning rate is 0.000005. The number of noun phrases $M$ is set to 10.

\begin{table*}[t]
\centering
\caption{Comparison with state-of-the-art approaches on CUHK-PEDES dataset. RR stands for Re-Ranking algorithm. Best results are in boldface.}
\label{tab:1}
\begin{tabular}{rccccc}
\hline
Method  & Scale & Top-1  & Top-5 & Top-10  & Total \\ \hline
GNA-RNN (CVPR'17)\cite{GNA-RNN} & \multirow{5}{*}{Global-Scale} & 19.05  & -  & 53.64   & - \\
PWM-ATH (WACV'18)\cite{PWM+ATH} &  & 27.14 & 49.45 & 61.02 & 137.61   \\
Dual-Path (TOMCCAP'20)\cite{Dual-Path} & & 44.40 & 66.26 & 75.07& 185.73 \\
CMPM+CMPC (ECCV'18)\cite{DCMP}    &      & 49.37& -  & 79.27    & -    \\
TIMAM (ICCV'19)\cite{TIMAM} & & 54.41   & 77.56  & 84.78 & 216.75    \\ \hline
PMA (AAAI'18)\cite{PMA} & \multirow{8}{*}{Multi-Scale} & 53.81  & 73.54 & 81.23 & 208.58 \\
MIA (TIP'19)\cite{MIA}  &  & 53.10  & 75.00  & 82.90& 211.00\\
ViTAA (ECCV'20)\cite{ViTAA}  &  & 55.97  & 75.84 & 83.52  & 215.33  \\
HGAN (MM'20)\cite{HGAN}   &  & 59.00& 79.49 & 86.62  & 225.11     \\
T-MRS(TCSVT'21)\cite{T-MRS} &   & 57.67   & 78.25   & 84.93  & 220.85 \\
MGEL (IJCAI'21)\cite{MGEL}  &  & 60.27  & 80.01 & 86.74 & 227.02 \\
SSAN (arXiv'21)\cite{SSAN}  &  & 61.37  & 80.15  & 86.73  & 228.25   \\
AXM-Net (arXiv'21)\cite{AXM-Net} & & 61.90   & 79.41  & 85.75& 227.06\\
DSSL (MM'21)\cite{DSSL}  &  & 59.98  & 80.41  & 87.56   & 227.95 \\
DSSL (MM'21)\cite{DSSL}+RR &  & 62.33 & 82.11& 88.01  & 232.45\\
TIPCB (Neurocomputing'22)\cite{TIPCB}    & & \textbf{63.63} & \textbf{82.82}     &\textbf{ 89.01} & \textbf{235.46} \\ \hline
NAFS (arXiv'21)\cite{NAFS}& \multirow{2}{*}{Adaptive Full-Scale} & 59.94 & 79.86 & 86.70  & 226.50 \\
NAFS (arXiv'21)\cite{NAFS}+RR  && 61.50& 81.19  & 87.51 & 230.20\\ \hline
\textbf{ACSA (Ours)} & \multirow{2}{*}{Asymmetric Cross-Scale} & 63.56 & 81.40 & 87.70 & 232.66 \\
\textbf{ACSA+RR} & & \textbf{68.67} & \textbf{85.61} & \textbf{90.66} & \textbf{244.94} \\ \hline
\end{tabular}
\end{table*}

\subsection{Comparison with State-of-the-Arts (SOTAs)}

We evaluated the proposed method by comparing to SOTA methods on both CUHK-PEDES and RSTPReid datasets. Experimental results are shown in Table \ref{tab:1} and \ref{tab:RSTPReid}, respectively. The state-of-the-art methods can be classified into three categories: 1) global-scale approaches (GNA-RNN \cite{GNA-RNN}, PWM+ATH \cite{PWM+ATH}, DCMP \cite{DCMP}, Dual-Path \cite{Dual-Path}, and TIMAM \cite{TIMAM}); 2) multi-scale approaches (PMA \cite{PMA}, MIA \cite{MIA}, ViTAA \cite{ViTAA}, HGAN \cite{HGAN}, T-MRS \cite{T-MRS}, MGEL \cite{MGEL}, SSAN \cite{SSAN}, AXM-Net \cite{AXM-Net}, DSSL \cite{DSSL}, and TIPCB \cite{TIPCB}) using both global and local scales; and 3) a full-scale approach (NAFS \cite{NAFS}), which uses global/local scales, and a cross scale.

From Table \ref{tab:1}, we make the following observations. First, the multi-scale approaches generally outperform the global-scale approaches. For instance, TIPCB \cite{TIPCB} achieves 63.63 at top-1 v.s. TIMAM \cite{TIMAM} with 54.41 at top-1. This proves the necessity of employing finer scale alignment. The full-scale approach, i.e., NAFS \cite{NAFS} also shows competitive results, suggesting that the alignment between different scales is beneficial to the task of TBPS. Comparing to these methods, our approach achieves similar performance to TIPCB \cite{TIPCB} in all evaluation indicators \textit{without re-ranking} (e.g., ACSA$\rightarrow$63.56 vs TIPCB \cite{TIPCB}$\rightarrow$63.63 on top-1). It is noteworthy that TIPCB \cite{TIPCB} uses multiple branches on top of different layers of the neural network to extract multi-scale visual features. This leads to high computational cost. In contrast, our approach achieves competitive performance using a simple network architecture.

To further improve the retrieval performance, we empirically employed a re-ranking algorithm \cite{NAFS} only in the testing phase. Obtained results show that re-ranking can noticeably improve the top-1 accuracy of our method by 5.11\%. As indicated by \cite{NAFS}, the re-ranking process requires both text-image and image-image retrieval. As such, the model not only needs to align the data of different modality, but also needs to fully leverage the information in each modality, especially in the image modality. In this experiment, we employed the self-attention mechanism to fully discover the informative priors in the image modality, and applied the cross-attention mechanism to effectively reduce the modal gap.

\begin{table}[t]
\centering
\caption{Comparison to the state-of-the-art method on RSTPReid dataset. Best results are in boldface.}
\label{tab:RSTPReid}
\begin{tabular}{lcccc}
\hline
Method  & Top-1 & Top-5 & Top-10    & Total \\ \hline
DSSL (ACM MM'21)\cite{DSSL} & 32.43 & 55.08 & 63.19  & 150.70  \\
\textbf{ACSA (Ours)} & \textbf{48.40} & \textbf{71.85} & \textbf{81.45} & \textbf{201.70} \\ \hline
\end{tabular}
\end{table}

Experimental results on RSTPReid dataset are reported in Table \ref{tab:RSTPReid}. RSTPReid is a newly introduced dataset for TBPS. As such, only DSSL \cite{DSSL} is validated on this dataset. Table \ref{tab:RSTPReid} shows that comparing to DSSL \cite{DSSL} our algorithm achieves performance gain by 15.97\%, 16.77\% and 18.26\% in terms of top-1, top-5 and top-10 respectively.

\subsection{Ablation Studies}
To thoroughly study the effectiveness of each module in our method, we conducted a series of experiments on CUHK-PEDES dataset, including different backbones for visual/textual representations, image partitioning and the effect of the proposed ACSA.

\subsubsection{Impact of Different Backbones}

In this experiment, we studied the impact of applying different backbones for the visual and textual domains. Specifically, we considered three image backbones, i.e., ResNet50 \cite{ResNet}, ViT \cite{ViT} and  Swin Transformer \cite{SwinTransformer}, and two text backbones, i.e., Bi-LSTM \cite{Bi-LSTM} and BERT \cite{BERT}. After extracting image and textual features using respective backbones, we simply performed a multi-scale alignment, and the cross-modal matching was performed by using two losses: CMPM \cite{DCMP} and CMPC \cite{DCMP}.

Experimental results are reported in Table \ref{tab:backbones}. When Bi-LSTM \cite{Bi-LSTM} is applied as the text backbone, we could observe that Swin Transformer \cite{SwinTransformer} brings 2.24\% performance gain on top-1 accuracy. When replacing Bi-LSTM \cite{Bi-LSTM} with BERT \cite{BERT}, the top-1 accuracy gain increases to 8.66\%. Similar results can be seen on top-5 and top-10. These results demonstrate the effectiveness of employing the Swin Transformer \cite{SwinTransformer} as the image backbone. Similarly, BERT \cite{BERT} outperforms Bi-LSTM \cite{Bi-LSTM} as the text backbone. This  demonstrates that the self-attention based network is effective in discovering the intra-modal information for feature representations. When we combine Swin Transformer \cite{SwinTransformer} and BERT \cite{BERT}, a notable improvement of 17.12\% is achieved on top-1 accuracy versus the combination of ResNet50 \cite{ResNet} and Bi-LSTM \cite{Bi-LSTM}. This empirically confirms the adoption of transformer-based networks in both domains. 

\begin{table}[t]
\centering
\caption{Impact of different backbones. Best results are in boldface.}
\label{tab:backbones}
\setlength{\tabcolsep}{1.5mm}
\begin{tabular}{rrcccc}
\hline
Image Model & Text Model & Top-1 & Top-5 & Top-10 & Total  \\ \hline
ResNet50 \cite{ResNet} & Bi-LSTM \cite{Bi-LSTM}  & 43.65 & 67.00 & 76.37  & 187.02 \\
ResNet50 \cite{ResNet} & BERT \cite{BERT} & 52.11 & 73.63 & 82.35  & 208.09 \\
ViT \cite{ViT} & Bi-LSTM \cite{Bi-LSTM} & 44.36 & 67.99 & 77.80 & 190.15\\
ViT \cite{ViT} & BERT \cite{BERT} & 59.61 & 79.19 & 85.56 & 224.36 \\
Swin Transformer \cite{SwinTransformer} & Bi-LSTM  \cite{Bi-LSTM} & 45.89 & 69.60 & 78.66  & 194.15 \\
Swin Transformer \cite{SwinTransformer} & BERT \cite{BERT} & \textbf{60.77} & \textbf{80.02} & \textbf{86.63} & \textbf{227.42} \\ \hline
\end{tabular}
\end{table}

\subsubsection{Different Image Partitioning Strategies}

\begin{table}[t]
\centering
\caption{Study on different local image feature extraction methods. Best results are in boldface.}
\label{tab:4}
\setlength{\tabcolsep}{1.5mm}
\begin{tabular}{ccccc}
\hline
Strategy & Top-1   & Top-5  & Top-10 & Total  \\ \hline
Six Slices \cite{MIA}  & 62.39 & 80.64 & 87.05  & 230.08\\
\textbf{Partition} & \textbf{63.56} & \textbf{81.40} & \textbf{87.70} & \textbf{232.66} \\ \hline
\end{tabular}
\end{table}

We employed a partitioning strategy to obtain the embedding of head, upper body, lower body and foot regions as local visual embeddings. Unlike the simple slicing strategy of directly dividing the global image embedding into six slices \cite{MIA}, our partitioning strategy brings no extra computational cost and ensures the consistency of local regions in higher layers of the network. We compared our partitioning strategy with the simple slicing method, and report the results in Table \ref{tab:4}. The results show that our partitioning strategy achieves better performance and effectively avoids splitting the same region into different slices.

\subsubsection{Impact of Asymmetric Cross-Scale Alignment}

\begin{table}[t]
\centering
\caption{Impact of different alignment relations. Best results are in boldface.}
\label{tab:5}
\setlength{\tabcolsep}{0.3mm}
\begin{tabular}{ccccccccc}
\hline
Relation & Image & Region & Text & Phrase & Top-1 & Top-5 & Top-10 & Total  \\ \hline
\uppercase\expandafter{\romannumeral1}   & ×     & × & ×  & × & 60.77 & 80.02 & 86.63  & 227.42 \\
\uppercase\expandafter{\romannumeral1} ,\uppercase\expandafter{\romannumeral2}   & × & \checkmark      & ×    & \checkmark      & 62.13 & 80.91 & 87.04  & 230.08 \\
\uppercase\expandafter{\romannumeral1} ,\uppercase\expandafter{\romannumeral2},\uppercase\expandafter{\romannumeral3},\uppercase\expandafter{\romannumeral4}  & \checkmark     & \checkmark      &\checkmark    & \checkmark      & 62.95 & 81.06 & 87.31  & 231.32 \\
\textbf{\uppercase\expandafter{\romannumeral1},\uppercase\expandafter{\romannumeral2},\uppercase\expandafter{\romannumeral3}} & \textbf{\checkmark} & \textbf{\checkmark} & \textbf{×} & \textbf{\checkmark} & \textbf{63.56} & \textbf{81.40} & \textbf{87.70} & \textbf{232.66} \\ \hline
\end{tabular}
\end{table}

To investigate the effectiveness of the proposed asymmetric cross-scale alignment, we implemented different scales of feature embedding in the cross-attention module, as shown in Table \ref{tab:5}. It should be noted that these experiments are based on global-level alignment, that is, the alignment in the cross-attention module is a supplement to global-level alignment. 

\textbf{Multi-Scale Alignment.} We only employed local embeddings for visual and text in the cross-attention module, i.e., region embeddings and noun phrase embeddings. It can be regarded as a multi-scale alignment approach that includes global-level and local-level alignments, i.e., the relations \uppercase\expandafter{\romannumeral1} and \uppercase\expandafter{\romannumeral2} in Fig. \ref{fig:alignment}. Table \ref{tab:5} shows that it improves the top-1 accuracy by 1.36\% in comparing to the global-level alignment. This proves the effectiveness of using finer-scale alignment.

\textbf{Adaptive Full-Scale Alignment.} Based on the above multi-scale alignment, we further considered the cross-scale alignment, i.e., region-text alignment and image-phrase alignment. That is, it includes relations \uppercase\expandafter{\romannumeral1}$ \sim $\uppercase\expandafter{\romannumeral4}. Specifically, we concatenated the region embeddings and global image embedding as the final visual embeddings. We also concatenated noun phrase embeddings and global text embedding as the final textual embeddings. Compared with the multi-scale alignment, e.g., relations \uppercase\expandafter{\romannumeral1}$ \sim $\uppercase\expandafter{\romannumeral4} , our method achieves better results by applying the cross-scale alignment.

\textbf{Asymmetric Cross-Scale Alignment.} Intuitively, we argue that region-text alignment in adaptive full-scale alignment is unnecessary. Thus, in the cross-attention module, we concatenated region embeddings and global image embedding as the final visual embeddings but only utilized the noun phrase embeddings as the final textual embeddings. That is, we considered the relations \uppercase\expandafter{\romannumeral1}$ \sim $\uppercase\expandafter{\romannumeral3}. It can be seen from Table \ref{tab:5} that this combination shows inferior performance, which indicates that region-text alignment is unnecessary. In fact, the entire texts rarely correspond to a specific region of a person image. Therefore, employing this alignment offers no benefits to the task of TBPS.

\subsubsection{Evaluation on Different Language Models}

In this experiment, we evaluated the proposed method using different language models, i.e., BERT \cite{BERT} and XLNet \cite{XLNet}. We considered the variations of our method and TIPCB \cite{TIPCB} by using the two language models. Specifically, we replaced the BERT \cite{BERT} in both the proposed method and TIPCB \cite{TIPCB} by using a recent language model XLNet \cite{XLNet}. Experimental results are reported in Table \ref{tab:language-models}. We have the following observations. Both our approach and TIPCB \cite{TIPCB} show better results than the variants of using XLNet \cite{XLNet}. This affirms the effectiveness of using BERT \cite{BERT} as the textual backbone. One possible reason is BERT \cite{BERT} adopts the random sampling, which produces more robust textual representations.

\begin{table*}[t]
\centering
\caption{Evaluation on different language models on CUHK-PEDES dataset. Best results are in boldface.}
\label{tab:language-models}
\begin{tabular}{rcccccc}
\hline
Method  & Visual Model & Language Model & Top-1  & Top-5 & Top-10 & Total \\ \hline
\multirow{2}{*}{TIPCB\cite{TIPCB}}& ResNet50 \cite{ResNet} & BERT\cite{BERT} & \textbf{63.22} & \textbf{82.79}   &\textbf{ 89.92} & \textbf{234.93} \\ 
& ResNet50 \cite{ResNet} & XLNet \cite{XLNet} & 58.41 & 79.78 &86.47 & 224.66 \\ 
\hline
\multirow{2}{*}{ACSA (Ours)} & Swin Transformer \cite{SwinTransformer} & BERT \cite{BERT} & \textbf{63.56} & \textbf{81.40} & \textbf{87.70} & \textbf{232.66} \\
 & Swin Transformer \cite{SwinTransformer} & XLNet \cite{XLNet} & 59.22 & 79.64 & 86.35 & 225.21 \\
 
\hline
\end{tabular}
\end{table*}

\subsection{Visualization}

\begin{figure*}[!t]
    \centering
    \includegraphics[height=7cm,width=18cm]{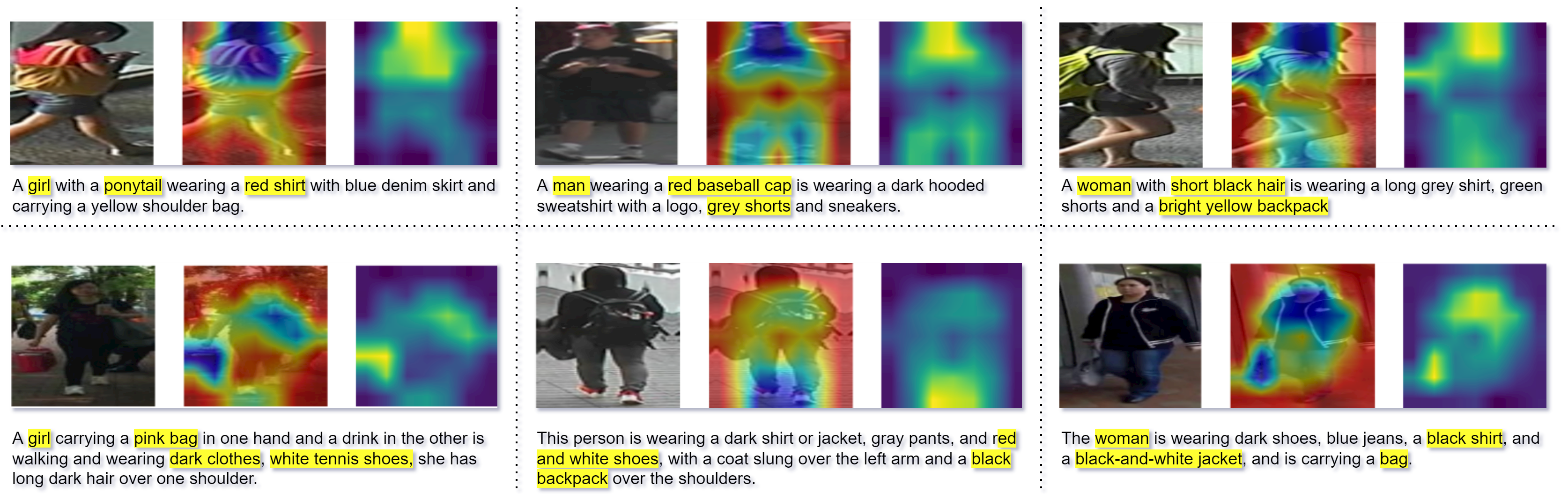}  
    \caption{Instance examples with saliency through the attention maps. For each group, the leftmost image is the raw pedestrian image, followed by the heat map of attention in the middle, and the highlighted areas in the right one represent the attentive saliency information.}
    \label{fig:attention_map}
\end{figure*}

\begin{figure}[t] 
	\centering  
	\includegraphics[height=12cm,width=8cm]{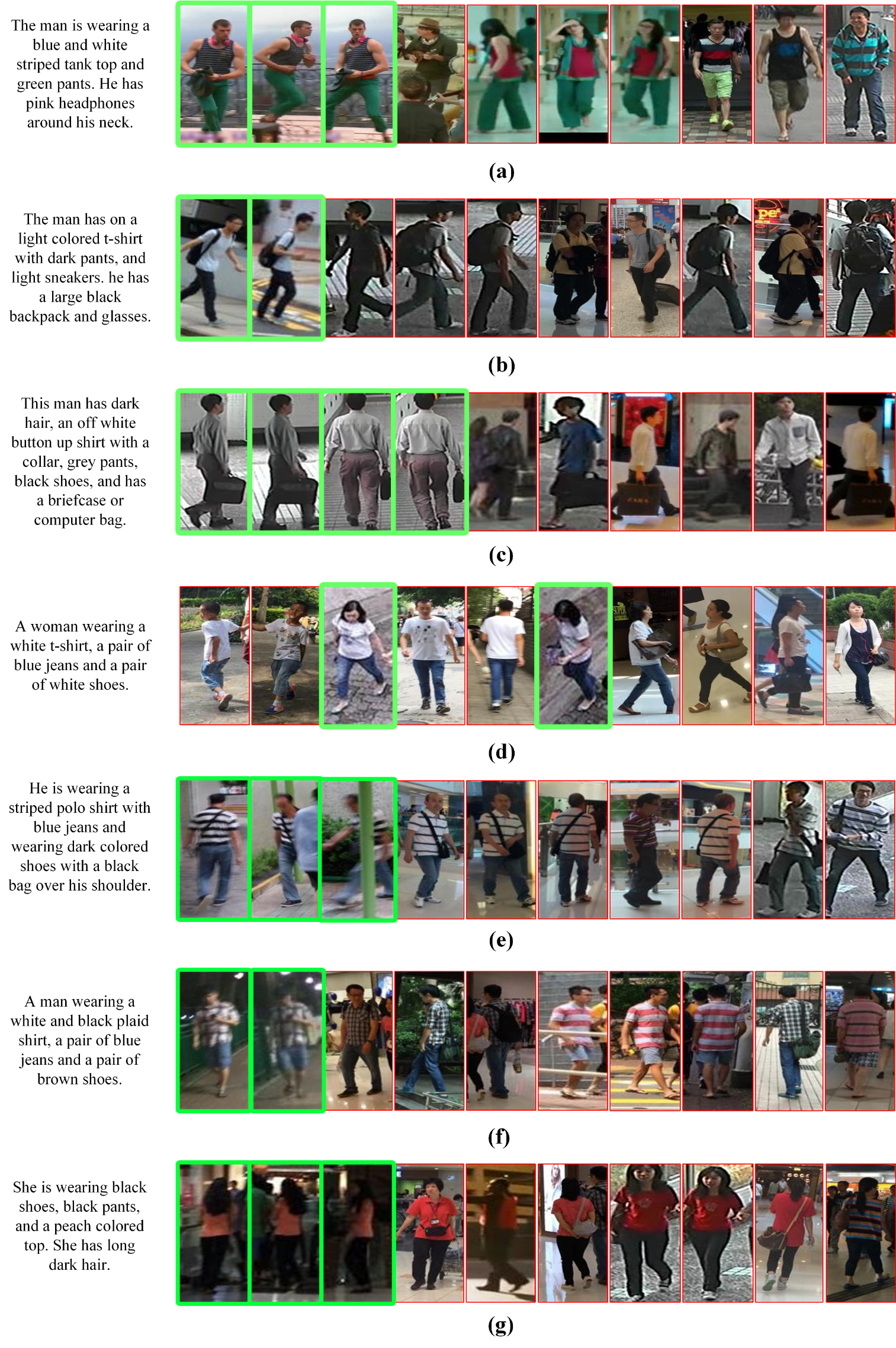}  
	\caption{Examples of top-10 retrieved images under various conditions (such as illuminations and night time) on the CUHK-PEDES dataset. Correct/incorrect images are marked by green/red rectangles.}
	\label{fig:retrieval_results}
\end{figure}

In Fig. \ref{fig:attention_map}, we show the saliency information in pedestrian images which can be learnt through the attention maps. In each group, we arrayed three images: the leftmost is the raw pedestrian images, the middle is the corresponding heat map based on the attention, and the right is the attentive saliency. We can see that the right image clearly shows the area where the saliency is located. In specific, our model can effectively ignore the background while focusing on its saliency on pedestrian images. In fact, the saliency information is mainly concentrated in four regions, i.e., head, upper body, lower body and the feet region, which verifies the rationale of our proposed partition strategy. Although there are some exceptions that some images may not meet this criterion, our partition strategy suits to human cognitive perception. Finally, it is noted that our model reveals a good correspondence between noun phrases and the saliency in the image. This demonstrates that our model is effective in aligning the cross-modal data. In Fig. \ref{fig:retrieval_results}, we further show retrieval results with pedestrian images captured under various conditions, such as illuminations and nighttime. The results show that most of the correct images appear in the first few ranked positions of the retrieval list. This proves that our model is robust against nuisance factors, and can be applied into various real-world conditions.

\section{Conclusion}
In this paper, we propose a transformer-based model for text-based person search by employing the Swin Transformer \cite{SwinTransformer} and BERT \cite{BERT} to extract multi-scale features from images and texts. To allow for fine-grained visual-text matching, we propose an Asymmetric Cross-Scale Alignment for adaptive cross-modal match, which consists of a global visual-text alignment, and an asymmetric cross-attention module for region/image-phrase alignments. Extensive experiments on CUHK-PEDES and RSTPReid datasets demonstrated the effectiveness and superiority of our approach.

\bibliographystyle{ieeetr}
\bibliography{IEEEabrv, mybibfile}

\begin{IEEEbiography}[{\includegraphics[width=1in,height=1.25in,clip,keepaspectratio]{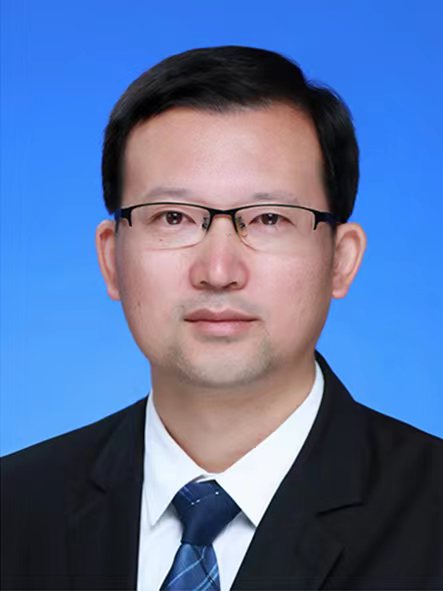}}] {Zhong Ji} received the Ph.D. degree in signal and information processing from Tianjin University, Tianjin, China, in 2008. He is currently a Professor with the School of Electrical and Information Engineering, Tianjin University. He has authored over 100 technical articles in refereed journals and proceedings, including IEEE TIP, IEEE TNNLS, IEEE TCYB, IEEE TCSVT, PR, CVPR, ICCV, ECCV, NeurIPS, AAAI, and IJCAI. His current research interests include zero/few-shot leanring, and cross-modal analysis.
\end{IEEEbiography}

\begin{IEEEbiography}[{\includegraphics[width=1in,height=1.25in,clip,keepaspectratio]{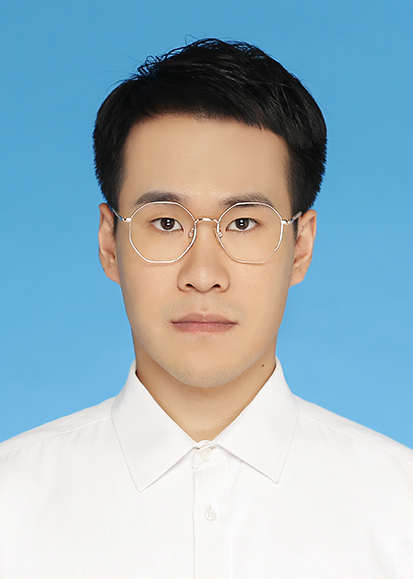}}] {Junhua Hu} received his B.S. degree in Electronic Information Engineering from Hebei University of Technology, Tianjin, China, in 2019. He is currently pursuing his M.S. degree in School of Electrical and Information Engineering, Tianjin University. His research interests include text-based person search and self-supervised learning.
\end{IEEEbiography}

\begin{IEEEbiography}[{\includegraphics[width=1in,height=1.25in,clip,keepaspectratio]{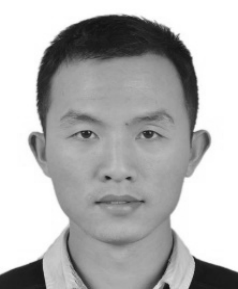}}] {Deyin Liu} received his B.E. and Ph.D degree from Zhengzhou University, China, in 2010 and 2021, respectively. He is currently working as a lecturer with School of Artificial Intelligence, Anhui University, China. His main research interests include optimization in computer vision, unsupervised learning and sparse representation learning.
\end{IEEEbiography}

\begin{IEEEbiography}[{\includegraphics[width=1in,height=1.25in,clip,keepaspectratio]{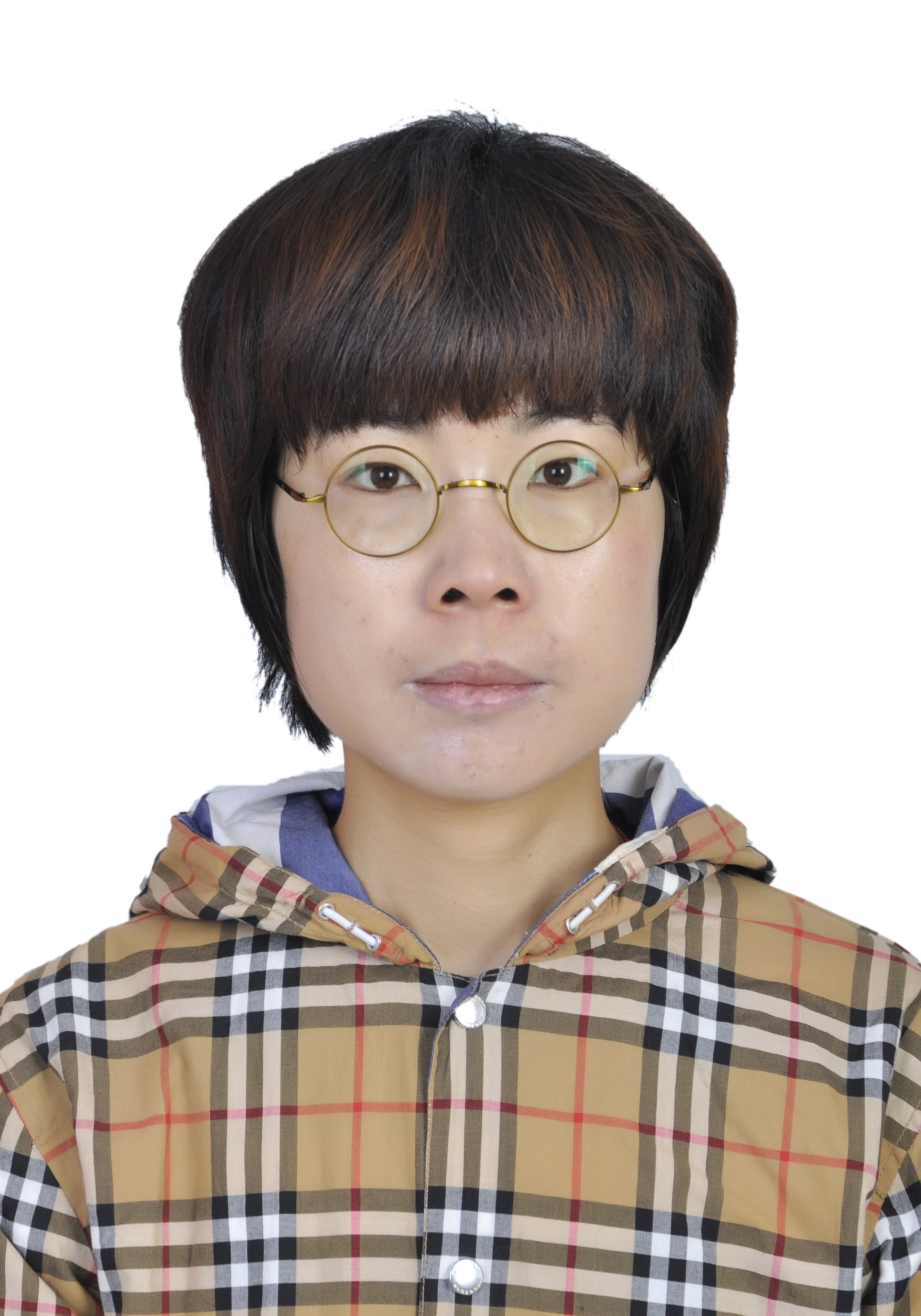}}]{Lin Yuanbo Wu} received a Ph.D. from The University of New South Wales, Australia in 2014. She is currently working as a senior lecturer (associate professor) with Department of Computer Science, Swansea University, UK. She was previously working in the University of Western Australia, Hefei University of Technology (China), the University of Queensland, and the University of Adelaide, Australia. Her research outcome are expounded with 60+ academic papers (including two book chapters) in premier journals and proceedings. She served as an Area Chair with ACM Multimedia 2022.
\end{IEEEbiography}

\begin{IEEEbiography}[{\includegraphics[width=1in,height=1.25in,clip,keepaspectratio]{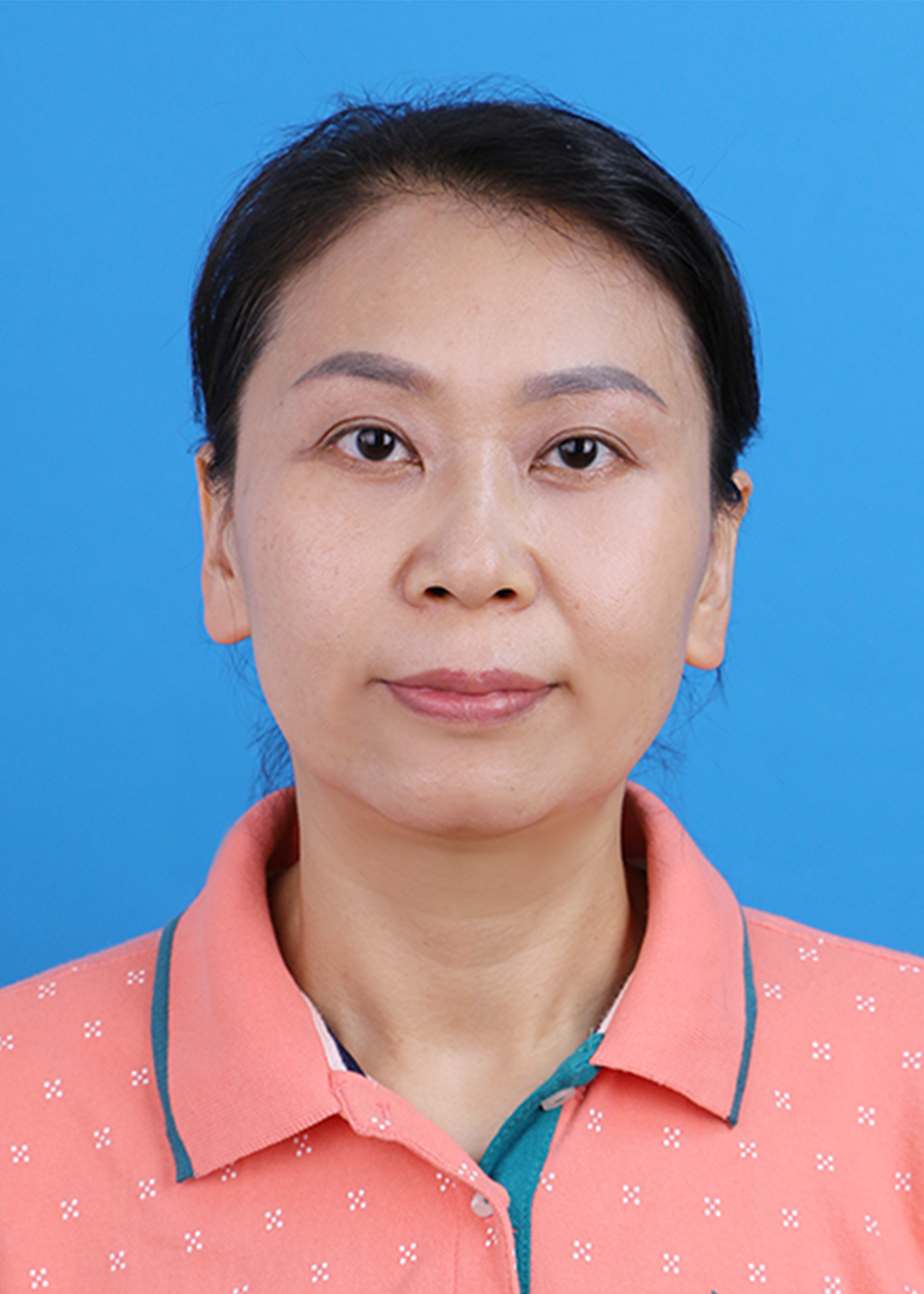}}]{Ye Zhao} received the M.S. degree in communication and information system from Harbin Engineering University, Harbin, China, in 2005 and the Ph.D. degree in signal and information processing from Hefei University of Technology, Hefei, China, in 2014.
She is an associate professor in School of Computer and Information, Hefei University of Technology. From 2016 to 2017, she was a visiting scholar in Computer Science department, University of Central Florida, USA. Her research interest includes Multimedia Analysis and Pattern Recognition. 
\end{IEEEbiography}

\end{document}